\def\paperlanguage{}
\pdfoutput=1 

\documentclass{tADR2e}

\usepackage{bm}
\usepackage{cite}
\include{preamble}

\newcommand{\ctext}[1]{\raise0.2ex\hbox{\textcircled{\scriptsize{#1}}}}

\title{\LARGE \textbf
  {
    \switchlanguage%
    {%
      Robotic Environmental State Recognition\\ with Pre-Trained Vision-Language Models\\ and Black-Box Optimization
    }%
    {%
      ブラックボックス最適化と事前学習済み大規模視覚-言語モデルに基づくロボットにおける環境状態認識
    }%
  }
}

\author{Kento Kawaharazuka$^{a}$$^{\ast}$, Yoshiki Obinata$^{a}$, Naoaki Kanazawa$^{a}$, Kei Okada$^{a}$, and Masayuki Inaba$^{a}$
    \thanks{$^\ast$Corresponding author. Email: kawaharazuka@jsk.imi.i.u-tokyo.ac.jp \vspace{6pt}}\\\vspace{6pt}  $^{a}${\em{The Department of Mechano-Informatics, Graduate School of Information Science and Technology, The University of Tokyo, 7-3-1 Hongo, Bunkyo-ku, Tokyo, Japan.}}
  }
\begin{document}

\jvol{00} \jnum{00} \jyear{2024} \jmonth{August}

\maketitle

\begin{abstract}
  \switchlanguage%
  {%
    In order for robots to autonomously navigate and operate in diverse environments, it is essential for them to recognize the state of their environment.
    On the other hand, the environmental state recognition has traditionally involved distinct methods tailored to each state to be recognized.
    In this study, we perform a unified environmental state recognition for robots through the spoken language with pre-trained large-scale vision-language models.
    We apply Visual Question Answering and Image-to-Text Retrieval, which are tasks of Vision-Language Models.
    We show that with our method, it is possible to recognize not only whether a room door is open/closed, but also whether a transparent door is open/closed and whether water is running in a sink, without training neural networks or manual programming.
    In addition, the recognition accuracy can be improved by selecting appropriate texts from the set of prepared texts based on black-box optimization.
    For each state recognition, only the text set and its weighting need to be changed, eliminating the need to prepare multiple different models and programs, and facilitating the management of source code and computer resource.
    We experimentally demonstrate the effectiveness of our method and apply it to the recognition behavior on a mobile robot, Fetch.
  }%
  {%
    ロボットが自律的に多様な環境を移動し活動するためには, その環境の状態を認識することが不可欠である.
    一方で, これらは人間が直接画像や点群をプログラムで処理して行う, アノテーションを行いデータセットを作成して学習するそれぞれの状態認識に対して, それに適した個別の手法が取られてきた.
    本研究では, ロボットのための環境状態認識を事前学習済みの大規模視覚-言語モデルにより言語を用いて統一的に行う.
    この際, 視覚-言語モデルのタスクであるVisual Question Answering (VQA)とImage-to-Text Retrieval (ITR)を応用する.
    本研究により, ニューラルネットワークの再学習や手動のプログラミング等無しに, ドアの開閉認識や電気のオンオフ認識だけでなく, これまで困難であった透明なドアの開閉や水の認識までも可能になることを示す.
    また, 用意した多数のテキスト集合の中からブラックボックス最適化に基づき適切なものを選ぶことで, よりその認識精度を向上させることが可能である.
    各状態認識について, 変化させるのはテキスト集合とその重みのみであり, 異なるモデルやプログラムを複数用意する必要がなく, コードやリソースの管理も容易である.
    実験から本研究の有効性を示すと同時に, 本手法を応用した認識行動を台車型ロボットFetch上に実装し動作を検証する.
  }%
\end{abstract}

\begin{keywords}
  Vision-Language Model, Black-Box Optimization, Environmental State Recognition
\end{keywords}

\section{Introduction}\label{sec:introduction}
\switchlanguage%
{%
  For robots that perform tasks such as daily life support, nursing care, and security, recognition of the surrounding environment is indispensable \cite{okada2006tool, saito2011subwaydemo}.
  For example, the robot must recognize whether a door is open, a light is on, water is running, a fire is burning, and so on.
  In order to change the robot's behavior based on the recognition results, state recognition is usually performed with discrete values of about two or three options.
  Until now, appropriate individual methods have been used for each state to be recognized, such as direct processing of images or point clouds by human programming \cite{chin1986recognition, quintana2018door}, creating a dataset with annotations and training neural networks \cite{li2020modifiedyolov3}, or detecting the state by installing new sensors \cite{chitta2010tactile, takahata2020coaxial}.
  However, these methods require us to manually program the process for each state recognition, to train neural networks one by one, and to increase the number of sensors installed.
  In addition, this will increase the number of programs and trained models needed for each state recognition, which will cause problems in management of source code and computer resource.
  To cope with these problems, a single program or model should be able to recognize multiple states.
}%
{%
  生活支援や介護, 警備などのタスクを行うロボットにおいて, 周囲の状態認識は欠かせない\cite{okada2006tool, saito2011subwaydemo}.
  これは例えば, ドアが開いているか, 電気がついているか, 水が出ているか, 火がついているかなどの状態認識である.
  認識結果に基づきロボットの行動を変化させるためには, 大抵2値や3値程度の離散的な状態認識が行われる.
  これまで, 人間が直接画像や点群をプログラムで処理して行う\cite{chin1986recognition, quintana2018door}, アノテーションを行いデータセットを作成して学習する\cite{li2020modifiedyolov3}, 新しく別のセンサを取り付け検知する\cite{chitta2010tactile, takahata2020coaxial}等, 認識したい個別の状態に対して, それに適した個別の手法が取られてきた.
  しかしこれには, 人間が一つ一つの状態認識に対して, その処理を手動でプログラミングしたり, NNを一つ一つ学習する必要があったり, 搭載するセンサ数を増やしたりする必要がある.
  また, これによって状態認識ごとのプログラムや学習モデルが増えていき, コードやリソースの管理に問題が生じる.
  これらの問題にに対処するには, ある一つのプログラムやモデルで, 複数の状態認識が可能となる必要がある.
}%

\begin{figure}[htb]
  \centering
  \includegraphics[width=0.8\columnwidth]{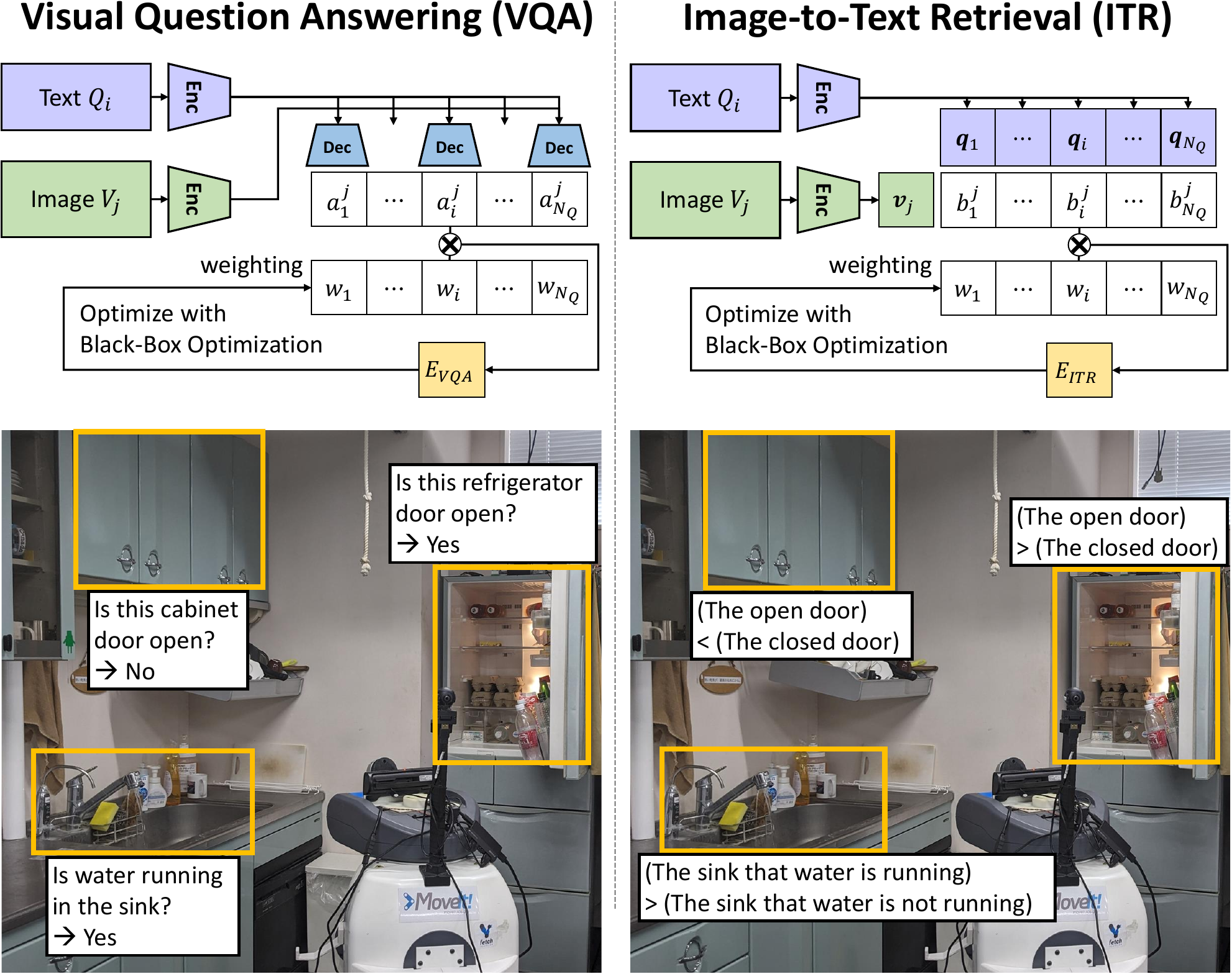}
  \caption{The concept of this study: for the robotic environmental state recognition, we use pre-trained vision-language models BLIP2 and OFA for Visual Question Answering (VQA), and CLIP and ImageBind for Image-to-Text Retrieval (ITR), with black-box optimization to optimize the weighting of prepared text prompts.}
  \label{figure:concept}
\end{figure}

\switchlanguage%
{%
  In this study, we propose a method to easily recognize various environmental states in a unified manner and through the spoken language (\figref{figure:concept}).
  In order to perform state recognition through the spoken language, we use pre-trained large-scale vision-language models (VLMs) \cite{li2022largemodels, li2023blip2, wang2022ofa, radford2021clip, girdhar2023imagebind}.
  Currently, VLMs are being used for map generation \cite{huang2022vlmaps, shafiullah2022clipfields}, scene understanding \cite{liu2023reflect, huang2023voxposer, shah2023navigation}, and feature extraction for behavior learning \cite{shridhar2021cliport}, in the context of robotics.
  In a few studies \cite{liu2023reflect, shah2023navigation}, environmental state recognition is implicitly performed, but there is no research discussing the characteristics and performance, and efforts toward improving the accuracy.
  VLMs are capable of performing a variety of tasks \cite{kawaharazuka2023ptvlm}, and we use Visual Question Answering (VQA), which returns answers to questions, or Image-to-Text Retrieval (ITR), which computes the correspondence between images and texts.
  In other words, state recognition is performed by inputting some text to the current image to be recognized or obtaining responses in the form of a sentence or degree of similarity.
  We show that various environmental states can be recognized only by changing the input text, and at the same time, we show how the performance varies from model to model.
  In addition, the recognition accuracy can be further improved by selecting appropriate texts from a large set of prepared texts.
  We show that more environmental states can be recognized by appropriately computing the weighting for each text based on black-box optimization.
  From the perspective of resource management, rather than conducting fine-tuning that requires individual model training and management, we leverage the versatility of the foundation model and the characteristics of language input.
  Note that this task is different from object detection or segmentation tasks in that it recognizes the state of the observed environment.

  This study makes it possible to recognize not only the open/closed state of room doors, but also various environmental states such as the open/closed state of transparent doors, whether water is running or not, and the cleanliness of the kitchen, through the spoken language.
  This environmental state recognition can be applied to the decision-making and if-else branching of actions in navigation, patrol, and life support tasks in robots.
  Since this study uses only pre-trained models, it does not require any training of neural networks or manual programming.
  For each state recognition, only the text and its weighting are changed, which eliminates the need to prepare multiple different models and programs, facilitating the management of source code and computer resource.
  Note that this study utilizes models of VQA and ITR independently; however, their state recognition is formulated within a unified framework.
  We will experimentally demonstrate the effectiveness of our method and apply it to the recognition behavior on a mobile robot, Fetch.

  The contributions of this study are as follows:
  \begin{itemize}
    \item Proposal of an environmental state recognition method using pre-trained large-scale vision-language models.
    \item Improvement of recognition accuracy through text weighting based on black-box optimization.
    \item Evaluation of the effectiveness, characteristics, and performance of the proposed method through experiments.
  \end{itemize}
}%
{%
  そこで本研究では, 多様な環境状態認識を, 統一的かつ言語を用いて容易に行う手法を提案する(\figref{figure:concept}).
  この言語を用いた状態認識を行うにあたり, 本研究では事前学習済みの大規模視覚-言語モデル\cite{li2022largemodels, radford2021clip, wang2022ofa, li2023blip2, girdhar2023imagebind}を用いる.
  現在, VLMsはロボットにおけるマップ生成\cite{huang2022vlmaps, shafiullah2022clipfields}, シーン理解\cite{liu2023reflect, huang2023voxposer}, 動作学習の特徴量\cite{shridhar2021cliport}などとして, 様々な形で使われ始めている\cite{kawaharazuka2024foundation}.
  一部の研究\cite{liu2023reflect}では環境状態認識が暗黙的に行われているが, これらの特性や性能, 精度向上に向けた取り組みについて議論した論文はない.
  大規模視覚-言語モデルは様々なタスクが可能であるが\cite{kawaharazuka2023ptvlm}, ここでは質問に対して答えを返すVisual Question Answering (VQA)と, 画像と言語の対応を計算するImage-to-Text Retrieval (ITR)の2つを応用する.
  つまり, 認識したい現在の画像に対して, 何らかのテキストを入力することで, 回答を文章または類似度という形で得ることで状態認識を行う.
  これにより, 様々な環境状態認識がテキストの変更のみで可能であることを示すと同時に, モデルごとにおける性能変化について示す.
  また, 用意した多数のテキスト集合の中から適切なものを選ぶことで, よりその認識精度を向上させることが可能である.
  各テキストに対する重み付けをブラックボックス最適化に基づいて適切に計算することで, より多くの環境状態認識が可能となることを示す.
  リソース管理の観点から, 個別のモデル学習と管理を要するfine-tuningは行わずに, 基盤モデルの汎用性と言語入力の特徴を活かした手法を提案する.
  なお, 本研究はこれまでの物体検出タスク等とは異なる, 観測した環境の状態を認識するタスクである点に注意されたい.

  本研究は, 言語を介することで, ドアの開閉だけでなく, これまで困難であった透明なドアの開閉や水の認識, 指標の曖昧な台所の綺麗さなど, 様々な状態認識が可能になる.
  この環境状態認識は, ロボットにおけるナビゲーションやパトロール, 生活支援などにおける, 移動可否の判断や行動の分岐に応用することができる.
  本研究は事前学習済みモデルのみ用いるため, ニューラルネットワークの再学習や手動のプログラミング等は一切必要ない, 非常にシンプルな手法と言える.
  各状態認識について, 変化させるのはテキストとその重みのみであり, 異なるモデルやプログラムを複数用意する必要がない点で優位性がある.
  なお, 本研究はVQAやITRのモデルを独立して用いているが, それら状態認識は統一的な枠組みで定式化されている.
  実験から本研究の有効性を示すと同時に, 本手法を応用した認識行動を台車型ロボットFetch上に実装し動作を検証する.

  本研究の貢献は以下の通りである:
  \begin{itemize}
    \item 事前学習済みの大規模視覚-言語モデルを用いた環境状態認識手法の提案
    \item ブラックボックス最適化に基づくテキストの重み付けによる認識精度の向上
    \item 実験を通した提案手法の有効性, その特徴, および性能の評価
  \end{itemize}
}%

\section{Robotic Environmental State Recognition with Pre-Trained Vision-Language Models and Black-Box Optimization} \label{sec:proposed}
\switchlanguage%
{%
  First, we describe large-scale vision-language models (VLMs), the tasks they can be used for, and the pre-trained models we actually use.
  Next, we describe a method for recognizing environmental states using these models, and finally, we describe a method for improving recognition accuracy by adjusting the weighting of texts based on black-box optimization.
}%
{%
  まず大規模視覚-言語モデルとそれらが可能なタスク, 実際の事前学習済みモデルについて述べる.
  次に, それらモデルを用いた環境状態認識手法について述べ, 最後にブラックボックス最適化に基づく重み調整により認識精度を高める方法について述べる.
}%

\subsection{Pre-Trained Vision-Language Models} \label{subsec:vlms}
\switchlanguage%
{%
  Various VLMs have been proposed so far.
  In terms of the tasks that VLMs are capable of, \cite{li2022largemodels} classifies them into four categories: Generation Task, Understanding Task, Retrieval Task, and Grounding Task.
  Generation Task includes Image Captioning (IC), which generates image captions, and Text-to-Image Generation (TIG), which generates images from language.
  Understanding Task includes Visual Question Answering (VQA), which answers questions about images, Visual Dialog (VD), which answers questions based on images and dialog history, Visual Reasoning (VR), which answers the reason in addition to VQA, and Visual Entailment (VE), which verifies the semantic validity of image-language pairs.
  Retrieval Task includes Image-to-Text Retrieval (ITR) and Text-to-Image Retrieval (TIR), which retrieve text or image from alternatives by calculating the correspondence between text and image.
  Grounding Task includes Phrase Grounding (PG) and Reference Expression Comprehension (REC), which extract the corresponding parts of an image from the language (PG and REC combined is expressed as Visual Grounding (VG)).
  Among these, tasks that directly output images like TIG, search for images like TIR, or extract bounding boxes in images like VG, are not suitable for state recognition.
  Tasks that handle dialog history and reason answering such as VD and VR, and tasks that directly output long sentences like IC, are also not suitable.
  On the other hand, VQA and ITR can be used for state recognition (VE is not considered since it is implicitly included in VQA).
  In other words, state recognition can be achieved by asking questions to the current image and obtaining ``Yes'' or ``No'' answers, or by obtaining the similarity between the current image and the prepared sentences.

  There are many pre-trained models that can perform VQA and ITR.
  As representatives, VQA models of BLIP2 \cite{li2023blip2} and OFA \cite{wang2022ofa}, and ITR models of CLIP \cite{radford2021clip} and ImageBind \cite{girdhar2023imagebind} are used in this study.
  BLIP2 is a model in which inputting an image $V$ and a question text $Q$, e.g. ``How many people are there?'', can generate an answer text $A$ such as ``two''.
  OFA is a similar model, but by learning multiple vision-language tasks at the same time, it has a high generalization capability that enables IC, TIG, VQA, VE, and VG in a single model.
  CLIP is a model that can calculate the cosine similarity between $\bm{v}$ and $\bm{q}$ vectorized from an image $V$ and a text $Q$, respectively.
  ImageBind is a model that can compute similarity not only for images and texts, but also for many other modalities including audio, depth images, heatmaps, and inertial sensors.
  Although the performance when using only OFA \cite{wang2022ofa} has been explored in \cite{kawaharazuka2023ofaga}, this study describes state recognition using VQA or ITR in a unified manner and discusses the differences among the models (the formulation is also different).
}%
{%
  大規模視覚-言語モデルには様々な形が提案されている.
  それらモデルが可能なタスクについて, \cite{li2022largemodels}ではGeneration Task, Understanding Task, Retrieval Task, Grounding Taskの4つに分類している.
  Generation Taskには, 画像のキャプションを生成するImage Captioning (IC)と, 言語から画像を生成するText-to-Image Generation (TIG)が含まれる.
  Understanding Taskには, 画像に関する質問に回答するVisual Question Answering (VQA), 画像と対話履歴から質問に回答するVisual Dialog (VD), VQAに加えてその理由を回答する必要のあるVisual Reasoning (VR), 画像と言語のペアに対してその意味的な妥当性を検証するVisual Entailment (VE)が含まれる.
  Retrieval Taskには, 画像と言語の対応を選択肢から検索するImage-to-Text Retrieval (ITR)とText-to-Image Retrieval (TIR)が含まれる.
  Grounding Taskには, 言語から画像中の対応する箇所のバウンディングボックスを抜き出すPhrase Grounding (PG)とReference Expression Comprehension (REC)が含まれる(これらはVisual Grounding (VG)として一つにまとめる).
  これらVLMsの中でも, TIGのように画像を直接出力する, TIRのように画像を検索する, VGのように画像中のバウンディングボックスを取り出すタスクは状態認識には向かない.
  また, VDやVRのように対話履歴や理由回答があるタスク, 直接長い文章が出力されるICも適切ではない.
  これらから, 主にVQAとITRの利用が状態認識には有効である(なお, VEはVQAに包含されるため直接は扱わない).
  つまり, 現在の画像に対して質問をすることで``yes''または``no''の回答を得たり, 現在の画像と用意した文章の類似度を得たりすることで, 状態認識が可能になる.

  このVQAとITRが可能な事前学習済みモデルは多数存在するが, その代表として, VQAについてはBLIP2 \cite{li2023blip2}とOFA \cite{wang2022ofa}, ITRについてはCLIP \cite{radford2021clip}とImageBind \cite{girdhar2023imagebind}を用いる.
  BLIP2は画像$V$と質問テキスト$Q$, 例えば``How many people are there?''を入力すれば, ``two''などの回答テキスト$A$を得ることが可能なモデルである.
  OFAも同様のモデルではあるが, 複数の視覚-言語タスクを同時に学習させることで, ICやTIG, VQA, VE, VG等が単一モデルで可能な高い汎化能力を得たモデルである.
  CLIPは画像$V$とテキスト$Q$を用意し, この画像とテキストをそれぞれ$\bm{v}$と$\bm{q}$にベクトル化, それらの間のコサイン類似度を計算することが可能なモデルである.
  ImageBindは, 画像やテキストのみでなく, 音声や深度画像, ヒートマップや慣性センサを含む多数のモーダルについて類似度が計算可能なモデルである.
}%

\subsection{Robotic Environmental State Recognition with Pre-Trained Vision-Language Models} \label{subsec:state-recognition}
\switchlanguage%
{%
  We describe a state recognition method for robots based on the pre-trained VLMs described in \secref{subsec:vlms}.

  First, we describe the state recognition by VQA.
  We input an appropriate question text $Q$ for an image $V$ and obtain a ``Yes'' or ``No'' answer $A$.
  For example, if the robot wants to recognize the open/closed state of a door, it can ask ``Is this door open?'' and if ``Yes'', it is open, and if ``No'', it is closed.
  In some cases, the answer $A$ may be neither ``Yes'' nor ``No'' such as ``this door is open'', in which case the answer is labeled as invalid.
  Since multiple answers $A$ can be obtained by adding random noise to $V$, the majority is used to obtain the answer.

  Next, we describe the state recognition by ITR.
  We prepare an appropriate text set $Q_{\{1, 2\}}$ for an image $V$, vectorize them, and compute the cosine similarity $\bm{v}^{T}\bm{q}_{\{1, 2\}}$.
  For example, if the robot wants to recognize the open/closed state of a door, let $Q_{1}$ be ``open door'' and $Q_{2}$ be ``closed door'', and if $\bm{v}^{T}\bm{q}_{1}\geq\bm{v}^{T}\bm{q}_{2}$, it is open, and if $\bm{v}^{T}\bm{q}_{1}<\bm{v}^{T}\bm{q}_{2}$, it is closed.
  Of course, if a threshold value is set, only one $Q$ is needed, and the door state can be recognized according to whether or not the cosine similarity exceeds the threshold value (not handled in this study).
  Since multiple similarities can be obtained by adding random noise to $V$, the average of these similarities is used to derive the answer.

  Here, the performance of the state recognition varies greatly for each $Q$.
  Therefore, the performance difference can be absorbed by preparing a large number of $Q$ in advance.
  In our experiments, we have prepared up to 80 $Q$ for each state to be recognized, with different articles, state expressions, and so on.
}%
{%
  \secref{subsec:vlms}で述べた事前学習済み大規模視覚-言語モデルを用いた, ロボットのための状態認識手法について述べる.

  まず, VQAによる状態認識について述べる.
  画像$V$に対して適切な質問テキスト$Q$を入力し, ``yes''または``no''の回答$A$を得る.
  例えばドアの開閉を認識したい場合, ``Is this door open?''と聞くことで, もし``yes''ならば開いている, ``no''ならば閉じていると認識することができる.
  場合によっては回答$A$が``this door is open''のような``yes''でも``no''でもない場合があるが, この際は回答を無効とする.
  $V$にランダムノイズを加えることで複数の回答$A$が得られるため, これらを統合して回答を導いても良い.

  次に, ITRの方法について述べる.
  画像$V$に対して適切なテキスト集合$Q_{\{1, 2\}}$を用意し, ベクトル化してコサイン類似度$\bm{v}^{T}\bm{q}_{\{1, 2\}}$を計算する.
  例えばドアの開閉を認識したい場合, $Q_{1}$を``open door'', $Q_{2}$を``closed door''とし, $\bm{v}^{T}\bm{q}_{1}\geq\bm{v}^{T}\bm{q}_{2}$ならば開いている, $\bm{v}^{T}\bm{q}_{1}<\bm{v}^{T}\bm{q}_{2}$ならば閉じていると認識することができる.
  もちろん閾値を設ければ, $Q$は一つとし, コサイン類似度が閾値を超えるか否かでドアの開閉を認識しても良い(本研究では扱わない).
  $V$にランダムノイズを加えることで複数の類似度が得られるため, これらを平均して回答を導いても良い.

  ここで, その状態認識性能は, $Q$ごとに大きく変化する.
  そのため, 事前に$Q$を多数用意することで, それら性能差を吸収することができる.
  本研究では, 一つの認識すべき状態について, 冠詞や状態表現などを変更した最大で80個の$Q$を用意し実験を行っている.
}%

\subsection{Robotic Environmental State Recognition Using Black-Box Optimization} \label{subsec:bb-optimization}
\switchlanguage%
{%
  In the method described in \secref{subsec:state-recognition}, $Q$ with both high and low performance are used uniformly.
  For states that are more difficult to be recognized, there is a possibility that the low performance $Q$ may adversely affect recognition ability.
  In addition, even for the same state recognition, the recognition performance for each $Q$ changes with changes in the angle of view, lighting, and so on.
  The recognition performance can be improved by finding an appropriate combination of $Q$ that enables correct state recognition under any condition.
  Therefore, we generate a highly accurate recognizer by appropriately selecting $Q$ by computing appropriate weights $w_{i}$ for a large number of $Q_i$ ($1 \leq i \leq N_Q$) through black-box optimization.
  We prepare a small dataset $D$ and evaluation function $E$, and perform unified optimization for state recognitions using VQA or ITR.

  First, we prepare the image $V_{j}$ ($1 \leq j \leq N_{V}$, where $N_{V}$ denotes the total number of images) and the corresponding correct responses $A^{j}_{D}$ as the dataset $D$.
  The angle of view and lighting are different for each image in the dataset.
  $A^{j}_{D}$ is a binary value of \{1, -1\}, and is labeled according to the state to be recognized, e.g., 1 is open and -1 is closed in the case of recognizing the open/closed state of doors.
  Note that the number of data with $A^{j}_{D} = 1$ and $A^{j}_{D} = -1$ are assumed to be the same.
  For each $Q_i$, let $Q^{1}_{i}$ be $Q$ for which the correct response is $1$ (e.g. ``Is this door open?'' or ``open door'') and $Q^{-1}_{i}$ be $Q$ for which the correct response is $-1$ (e.g. ``Is this door closed'' or ``closed door''), and the number of $Q^{1}_{i}$ and $Q^{-1}_{i}$ to be prepared is the same.

  Next, for each weight $w_{i}$ ($1 \leq i \leq N_{Q}$, $0 \leq w_i \leq 1$) of the texts, we set the evaluation function $E$ to be maximized based on black-box optimization.
  To compute the evaluation function, it is necessary to compute the percentage of correct responses $a^{j}_{i}$ for VQA or the degree of similarity $b^{j}_{i}$ for ITR, regarding each text $Q_i$ for each image $V_{j}$.
  In this study, for each image $V_{j}$, we augment the data by generating $N_{rand}$ images by RGBShift, which adds a random value from a uniform distribution within the range [-0.1, 0.1] to each RGB value (we set $N_{rand}=5$).
  This is intended to enhance robustness against changes in lighting and camera conditions.
  Note that there are various other augmentation methods besides RGBShift, but this study does not focus extensively on them.
  For VQA, the correct response rate $a^{j}_{i}$ for these $N_{rand}$ images and $Q_i$ is calculated as follows,
  \begin{align}
    a^{j}_{i} = \frac{N_{correct}}{N_{correct} + N_{wrong}}
  \end{align}
  where $N_{correct}$ is the number of correct responses, $N_{wrong}$ is the number of wrong responses, and invalid responses are ignored.
  If all the responses are invalid, we set $a^{j}_{i}=0$.
  For ITR, the average $b^{j}_{i}$ of the similarity between these $N_{rand}$ images and $Q_i$ is calculated as follows,
  \begin{align}
    b^{j}_{i} = \frac{1}{N_{rand}}\sum^{N_{rand}}_{k} \bm{v}_{j, k}^{T}\bm{q}_{i}
  \end{align}
  where $\bm{v}_{j, k}$ denotes the feature vector corresponding to the $k$-th ($1 \leq k \leq N_{rand}$) image of $V_{j}$ with random noise.
  By using these values, we compute the evaluation function $E$.
  For VQA, the evaluation function $E_{VQA}$ is set as follows,
  \begin{align}
    a^{j}_{w} &= \sum^{N_{Q}}_{i}w_{i}a^{j}_{i}/\sum^{N_{Q}}_{i}w_{i} \\
    E_{VQA} &= \sum^{N_{V}}_{j}\textrm{bool}(a^{j}_{w} > 0.5) + \alpha\sum^{N_{V}}_{j}a^{j}_{w} \label{eq:e-vqa}
  \end{align}
  where $\textrm{bool}(cond)$ is a function that returns 1 when $cond$ is satisfied and 0 otherwise, and $\alpha$ is a coefficient ($\alpha=0.01$ in this study).
  The image is correctly recognized when the weighted correct response rate $a^{j}_{w}$ exceeds $0.5$.
  In other words, the optimization is performed to maximize the number of correct responses, and then the sum of the correct response rate, for each data in $D$.
  For ITR, the evaluation function $E_{ITR}$ is set as follows,
  \begin{align}
    b^{j}_{w} &= A^{j}_{D}\sum^{N_{Q}}_{i}p_{i}w_{i}a^{j}_{i}/\sum^{N_{Q}}_{i}w_{i} \label{eq:e-itr-1}\\
    E_{ITR} &= \sum^{N_{V}}_{j}\textrm{bool}(b^{j}_{w} > 0.0) + \beta\sum^{N_{V}}_{j}b^{j}_{w} \label{eq:e-itr}
  \end{align}
  where $p_{i}$ is a variable that returns $1$ for $Q^{1}_{i}$ and $-1$ for $Q^{-1}_{i}$, and $\beta$ is a coefficient ($\beta=0.01$ in this study).
  For example, if $Q_1$ is ``open door'', $Q_2$ is ``closed door'', and $w_{\{1, 2\}}=1.0$, $\sum^{N_{Q}}_{i}p_{i}w_{i}a^{j}_{i}$ in \equref{eq:e-itr-1} is $ a^{j}_{1}-a^{j}_{2}$.
  For images with $A^{j}_{D}=1$, similarity $a^{j}_{1}$ with ``open door'' should exceed $a^{j}_{2}$ with ``closed door'', and vice versa, as expressed by \equref{eq:e-itr}.
  As in VQA, the optimization is performed to maximize the number of correct responses, and then the sum of the correct response rate, for each data in $D$.

  Finally, black-box optimization is performed.
  In this study, we apply a general genetic algorithm using DEAP \cite{fortin2012deap}.
  The gene sequence represented by $w_i$ is optimized based on the maximization of $E$.
  Here, a blend crossover will be applied with a probability of 50\%, and a Gaussian mutation with mean $0$ and variance $0.1$ will be applied with a probability of 20\%.
  Individuals are selected by the function selTournament, where the tournament size is set to 5.
  For individual selection, the best individual among a tournament size (set to $5$ in this study) of randomly chosen individuals is selected.
  The number of individuals is set to 300, and the number of generations is set to 1000.
  Note that the optimization process takes approximately 60 seconds.
}%
{%
  \secref{subsec:state-recognition}の方法では, 性能の高い$Q$も性能の低い$Q$も一様に利用している.
  ゆえに, より認識の難しい状態については, 性能の低い$Q$が認識に悪影響を及ぼす可能性がある.
  また, 同じ状態認識でも, 画像の画角や照明等が変化すると, 各$Q$について認識の得意不得意が変化する.
  そのため, どのような状況であっても正しく状態認識可能な, 適切な$Q$の組み合わせを得ることで, その認識性能を向上させることができる.
  そこで本研究では, 用意した多数の$Q_i$ ($1 \leq i \leq N_Q$)について, それらに関する適切な重み$w_{i}$をブラックボックス最適化により計算することで, $Q$を適切に取捨選択し精度の高い認識器を生成する.
  少数のデータセット$D$と評価関数$E$を用意し, VQAとITRを用いた状態認識に対して, 統一的に最適化を行う.

  まず, データセット$D$として, 画像$V_{j}$ ($1 \leq j \leq N_{V}$. $N_{V}$は画像の枚数を表す)と, それに対応する正しい回答$A^{j}_{D}$を用意する.
  なお, データセット内の各画像について, その画角や照明等は異なる.
  $A^{j}_{D}$は \{1, -1\}の二値であり, 認識したい状態に応じて, 例えばドアの開閉であれば, 1を開いた状態, -1を閉じた状態とラベル付けする.
  なお, $A^{j}_{D} = 1$であるデータと$A^{j}_{D} = -1$であるデータの枚数は同じとする.
  また, 用意した各$Q_i$については, 正しい回答が$1$となる$Q$を$Q^{1}_{i}$ (例えば ``Is this door open?''や``open door''), $-1$となる$Q$を$Q^{-1}_{i}$ (例えば ``Is this door closed?''や``closed door'')と表現し, 用意する$Q^{1}_{i}$と$Q^{-1}_{i}$の比率は同じとする.

  次に, 各テキストに対する重み$w_{i}$ ($1 \leq i \leq N_{Q}$, $0 \leq w_i \leq 1$)について, ブラックボックス最適化に基づき最大化する評価関数$E$を設定する.
  評価関数を計算するにあたり, 各画像$V_{j}$に対する各テキスト$Q_i$の正答率$a^{j}_{i}$ (VQAについて), または類似度$b^{j}_{i}$ (ITRについて)を計算する必要がある.
  本研究では各画像$V_{j}$について, RGBそれぞれの値に[-0.1, 0.1]の範囲内の一様分布からランダムに選ばれた値を足し込むRGBShiftによって$N_{rand}$個の画像を生成しデータを拡張する(本研究では$N_{rand}=5$とする).
  これは照明やカメラの変化等に対するロバスト性を高めるためのものです.
  RGBShift以外にも多様な拡張方法がありますが, 本研究ではここに大きな焦点は当てていません.
  VQAについては, これら$N_{rand}$個の画像と$Q_i$をモデルに入力した際の正答率$a^{j}_{i}$を以下のように計算する.
  \begin{align}
    a^{j}_{i} = \frac{N_{correct}}{N_{correct} + N_{wrong}}
  \end{align}
  ここで, $N_{correct}$は正答数, $N_{wrong}$は間違った回答数とし, 無効な回答は無視する.
  また, 全ての回答が無効な場合は$a^{j}_{i}=0$とする.
  ITRについては, これら$N_{rand}$個の画像と$Q_i$をモデルに入力した際の類似度の平均$b^{j}_{i}$を以下のように計算する.
  \begin{align}
    b^{j}_{i} = \frac{1}{N_{rand}}\sum^{N_{rand}}_{k} \bm{v}_{j, k}^{T}\bm{q}_{i}
  \end{align}
  ここで, $\bm{v}_{j, k}$は$V_{j}$にランダムノイズを加えた$k$番目($1 \leq k \leq N_{rand}$)の画像に対応する特徴ベクトルを表す.
  これらを用いて, 評価関数$E$を計算する.
  VQAについては以下のように評価関数$E_{VQA}$を設定する.
  \begin{align}
    a^{j}_{w} &= \sum^{N_{Q}}_{i}w_{i}a^{j}_{i}/\sum^{N_{Q}}_{i}w_{i} \\
    E_{VQA} &= \sum^{N_{V}}_{j}\textrm{bool}(a^{j}_{w} > 0.5) + \alpha\sum^{N_{V}}_{j}a^{j}_{w} \label{eq:e-vqa}
  \end{align}
  ここで, $\textrm{bool}(\cdot)$は, 条件$\cdot$を満たすとき1, それ以外のとき0を返す関数, $\alpha$は重みの係数を表す(本研究では$\alpha=0.01$とした).
  重み付けした正答率$a^{j}_{w}$が$0.5$を超えたとき, その画像は正しく認識可能である. 
  つまり, データセット$D$に関する正解データ数, それぞれのデータに関する正解率の和を順に最大化するように最適化を行う.
  ITRについては以下のように評価関数$E_{ITR}$を設定する.
  \begin{align}
    b^{j}_{w} &= A^{j}_{D}\sum^{N_{Q}}_{i}p_{i}w_{i}a^{j}_{i}/\sum^{N_{Q}}_{i}w_{i} \label{eq:e-itr-1}\\
    E_{ITR} &= \sum^{N_{V}}_{j}\textrm{bool}(b^{j}_{w} > 0.0) + \beta\sum^{N_{V}}_{j}b^{j}_{w} \label{eq:e-itr}
  \end{align}
  ここで, $p_{i}$は, $Q^{1}_{i}$に対して$1$を, $Q^{-1}_{i}$に対して$-1$を返す変数, $\beta$は重みの係数を表す(本研究では$\beta=0.01$とした).
  例えば$Q_1$が``open door'', $Q_2$が``closed door'', $w_{\{1, 2\}}=1.0$とすると, \equref{eq:e-itr-1}における$\sum^{N_{Q}}_{i}p_{i}w_{i}a^{j}_{i}$の部分は$a^{j}_{1}-a^{j}_{2}$となる.
  $A^{j}_{D}=1$な画像に関しては, ``open door''との類似度$a^{j}_{1}$は``closed door''との類似度$a^{j}_{2}$を上回るべきであり, その逆もまた然りということを\equref{eq:e-itr}は表現している.
  VQAと同様に, データセット$D$に関する正解データ数, それぞれのデータに関する正解率の和を順に最大化するように最適化を行う.

  最後に, ブラックボックス最適化を行う.
  本研究ではアルゴリズムとして, Deap \cite{fortin2012deap}を用いた遺伝的アルゴリズムを適用する.
  $w_i$により表現された遺伝子配列を$E$の最大化に基づき最適化する.
  この際, 50\%の確率でblend crossover, 20\%の確率で平均0, 分散0.1のgaussian mutationを行う.
  個体選択は関数selTournamentで行い, トーナメントサイズを5, 個体数を300, 世代数を1000とする.
  なお, 最適化の時間は約60秒である.
}%

\begin{figure}[htb]
  \centering
  \includegraphics[width=1.0\columnwidth]{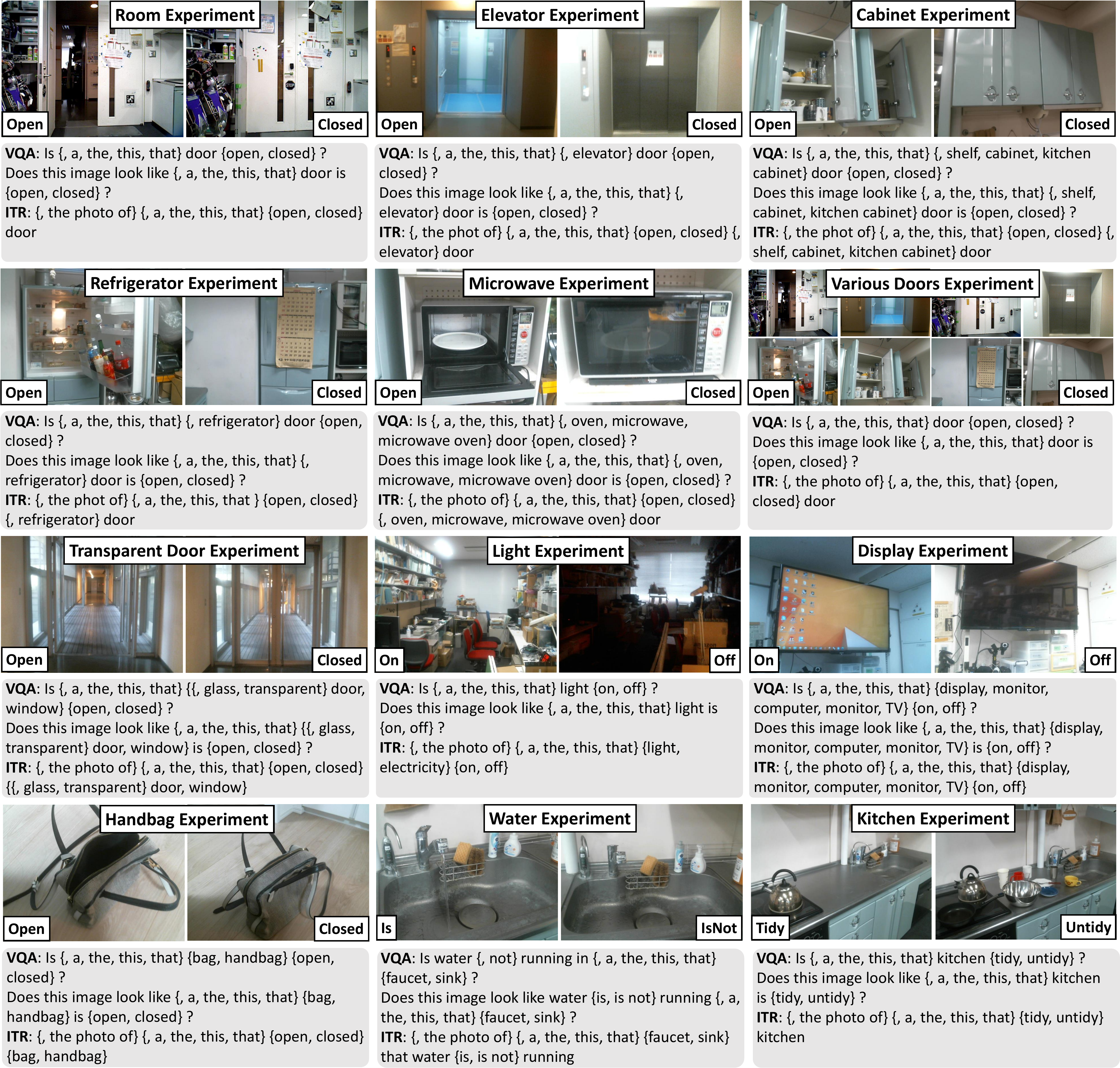}
  \caption{The set of text prompts and representative images for Room, Elevator, Cabinet, Refrigerator, Microwave, Various Doors, Transparent Door, Light, Display, Handbag, Water, and Kitchen experiments.}
  \label{figure:experiment}
\end{figure}

\section{Experiment} \label{subsec:experiments}
\switchlanguage%
{%
  Some of the images used in the state recognition experiments and the prepared text combinations are shown in \figref{figure:experiment}.
  Specifically, we conduct experiments to recognize the open/closed state of a room door, an elevator door, a cabinet door, a refrigerator door, a microwave oven door, various doors that are a combination of the above five doors, and a transparent door.
  Additionally, experiments were conducted to recognize the on/off state of lights and displays, the open/closed state of bags, whether water is running from a faucet or not, and the cleanliness of a kitchen.
  As a dataset $D_{opt}$ for optimization, we prepared 20 pictures for each experiment, e.g., open/closed doors (10 pictures each) and a faucet that water is running from or not (10 pictures each), at various angles of view.
  For ``Various Doors'' dataset, 100 pictures are prepared, 20 pictures for each of the five doors.
  As a dataset $D_{eval}$ for evaluation, we prepared the same number of data that is different from $D_{opt}$.
  $D_{eval}$ may have different lighting conditions from $D_{opt}$ because it is captured at different times, not just different angles of view.
  As for the text $Q$, we have prepared at most 80 $Q$ for each state to be recognized in our experiments.
  For both VQA and ITR, we prepared a large number of $Q$ by changing articles, state expressions, words, and question/expression forms.
  For articles, we use ``a'', ``the'', ``this'', ``that'', and no article.
  For the state expressions, antonyms such as ``open/closed'' and ``on/off'' are used (synonyms are also used).
  For words, synonyms such as ``glass door'', ``transparent door'', ``cabinet door'' and ``kitchen shelf door'' are used.
  For question/expression forms, we use changes in question forms such as ``Is the door open?'' and ``Does this image look like the door is open?'' for VQA, and changes in expression forms such as ``the open door'' and ``the photo of the open door'' for ITR.

  In this study, we perform comparative experiments using two models for VQA (BLIP2 and OFA) and two models for ITR (CLIP and ImageBind), for a total of four models.
  For each model, we evaluate the performance using both $D_{opt}$ and $D_{eval}$ datasets in three settings, \textbf{OPT}, \textbf{ONE}, and \textbf{ALL}.
  \textbf{OPT} is the result of applying the black-box optimization in this study.
  \textbf{ONE} is the result when only one $Q$ that maximizes $E$ is selected among the prepared $Q$ (since both $Q^{1}$ and $Q^{-1}$ must exist in the case of ITR, two $Q$ with a pair of state expressions are selected).
  \textbf{ALL} is the result when all the prepared $Q$ are used equally.
  For these settings, we compare the rate of correct state recognition.
  Note that the state recognition discussed here is often possible with an accuracy close to 100\%, depending on the method and settings, if human annotation, point cloud processing, and dedicated sensors are skillfully utilized.
  On the other hand, the important point of this study is that there is no need to train neural networks or manually program the process, and only the text and its weighting need to be provided for a single model, making it easy to manage source code and computer resource.
  Also, since this study involves optimizing the text weighting by collecting a small number of images, there is no validation data.
  In other words, optimization is performed using $D_{opt}$, and the performance is evaluated using both $D_{opt}$ and $D_{eval}$.
}%
{%
  本研究の状態認識実験に用いた画像の一部とそのテキストの組み合わせを\figref{figure:experiment}に示す.
  具体的には, 標準的なドア・エレベータのドア・棚のドア・冷蔵庫のドア・レンジのドア・上記5つを組み合わせた様々なドア・透明なドアの開閉, 明かり・ディスプレイのオンオフ, カバンの開閉, 蛇口からの出水, キッチンの綺麗さを認識する実験を行う.
  最適化用データセット$D_{opt}$として, 例えばドアについては開いている状態と閉じている状態を10枚ずつの計20枚, 水については蛇口から出ている状態と出ていない状態を10枚ずつの計20枚を様々な画角で用意した.
  なお, ``Various Doors''については5つのドアについて20枚ずつの計100枚のデータを用意している.
  評価用データセット$D_{eval}$として, $D_{opt}$とは異なる同じ枚数のデータを用意した.
  $D_{eval}$は$D_{opt}$とは別の時間帯で撮影しているため, 画角だけでなく照明条件も異なる場合がある.
  また, テキスト$Q$については, 一つの認識すべき状態について最大で80個の$Q$を用意し実験を行っている.
  VQAとITRの両者について, 冠詞・状態表現・単語・質問/表現形式を変更することで多数の$Q$を用意している.
  冠詞については, 例えば``a''や``the'', ``this''や``that'', 無冠詞の5種類を用いる.
  状態表現については, 例えば ``open''や``closed'', ``on''や``off''のような対義語を用いる(類義語の使用も可能である).
  単語については, 例えば, ``glass door''や``transparent door', ``cabinet door''や``shelf door''のような同義語を用いる.
  質問/表現形式については, 例えば``Is the door open?''や``Does this image look like the door is open?''の質問形式の変化, ``the open door''や``the photo of the open door''のような表現形式の変化を用いる.

  比較実験について述べる.
  本研究では, VQAについて2つのモデル(BLIP2とOFA), ITRについて2つのモデル(CLIPとImageBind)の, 計4つのモデルを用いて比較実験を行う.
  また, それぞれのモデルについて, \textbf{OPT}, \textbf{ONE}, \textbf{ALL}の3つの設定で, $D_{opt}$と$D_{eval}$の両者のデータセットについて評価する.
  \textbf{OPT}は本研究におけるブラックボックス最適化を適用した際の結果である.
  \textbf{ONE}は, 用意した$Q$の中から$E$を最大化する最善の$Q$を一つだけ選択した際の結果である(ITRの場合は$Q^{1}$と$Q^{-1}$の両者が存在する必要があるため, 状態表現が対となる二つの$Q$を選択している).
  \textbf{ALL}は, 用意した$Q$を全て均等に使った際の結果である.
  これらの設定について, 状態認識の正答率を比較する.
  なお, ここで扱う状態認識は人間によるアノテーションや点群処理, 専用のセンサを巧みに駆使すれば, 方法や設定次第では100\%近い高い精度の認識が可能な場合も多い.
  一方, 本研究の重要な点は, ニューラルネットワークの再学習や手動のプログラミング等がなく, かつ単一のモデルでテキストとそれらの重みのみ用意すれば良いためリソース管理が容易な点である.
  また, 本研究は少数枚の画像を集めて各textのweightを最適化するという設定のため, validation dataはありません.
  つまり, $D_{opt}$を用いて最適化を行い, その性能を$D_{opt}$と$D_{eval}$の両者で評価しています.
}%

\subsection{State Recognition Experiment}
\switchlanguage%
{%
  The results of the state recognition experiment are shown in \tabref{table:experiment}.
  The percentage of correct responses when applying the four models to the 12 states to be recognized are shown in \figref{figure:experiment}.
  The last two rows of \tabref{table:experiment} show the average and standard deviation of the correct rates.

  First, we discuss the results of \textbf{ALL}.
  Since \textbf{ALL} does not require any optimization, if all states can be recognized by this setting, state recognition will be much easier.
  VQA(OFA) achieves almost 100\% correct responses for both $D_{opt}$ and $D_{eval}$ in the state recognition of Room, Elevator, Cabinet, and Kitchen.
  In addition, ITR(ImageBind) achieves almost 100\% correct responses for the state recognition of Cabinet, Refrigerator, Microwave, HandBag, and Kitchen.
  In other words, it is possible to recognize simple states by merely describing the state to be recognized in the spoken language.
  On the other hand, the correct response for the state recognition of Transparent Door, Light, Display, and Water are not high with any model.
  Also, the performance of VQA(BLIP2) and ITR(CLIP) is lower than that of VQA(OFA) and ITR(ImageBind) in most cases.

  Next, we discuss the results of \textbf{OPT} with optimization.
  The performance of \textbf{OPT} is higher than that of \textbf{ALL} in most cases.
  For Light, Display, and Water, which are difficult to achieve high correct response rates with \textbf{ALL}, VQA(OFA) and ITR(ImageBind) achieve nearly 100\% correct responses.
  For Various Doors, which handles five types of doors with the same text, more than 90\% correct responses are achieved.
  Note that the performance of \textbf{ONE}, which uses only one $Q$, is worse than that of \textbf{ALL}.

  In terms of the average percentage of correct responses, VQA(OFA) and ITR(ImageBind) have similar results: \textbf{OPT} with optimization achieves about 95\%, and even \textbf{ALL} without optimization achieves more than 80\%.
  On the other hand, the performance of VQA(BLIP2) and ITR(CLIP) with \textbf{ALL} is much lower, with about 60\%.
  Moreover, the performance of ITR(CLIP) with \textbf{OPT} is only as good as that of VQA(OFA) and ITR(ImageBind) with \textbf{ALL}.
  We can see that for all models, the variance of the correct response rate for each recognized state is reduced by the optimization.
  In particular, the variance of the correct response rate of \textbf{OPT} in ITR(ImageBind) is much smaller than that of other models.
}%
{%
  本研究の結果を\tabref{table:experiment}に示す.
  \figref{figure:experiment}に示した12の状態認識について, 4つのモデルを適用した実験結果を示しており, 状態認識の正答率をパーセンテージで表している.
  \tabref{table:experiment}の最後の行には, 正答率の平均値と標準偏差を示している.

  まず, 最適化を用いない\textbf{ALL}について述べる.
  \textbf{ALL}は, 一切の最適化を必要としないため, もしこれで全ての状態認識が可能であるのであれば, これまでとは比較にならないほど容易に状態認識が可能となる.
  VQA(OFA)では, DoorやElevator, CabinetやKitchenの状態認識において, $D_{opt}$, $D_{eval}$ともに, ほぼ100\%の正答率を達成している.
  また, ITR(ImageBind)では, Cabinet, Refrigerator, Microwave, HandBag, Kitchenの状態認識について, 同様に100\%近い正答率を達成している.
  つまり, 簡単な状態認識であれば, 認識したい状態を言語で記述するのみで, 状態認識が可能であることがわかる.
  一方で, Transparent DoorやLight, DisplayやWaterのような状態認識については, どのモデルでも正答率は高くない.
  また, 同じVQAやITRでも, 別のモデルであるBLIP2やCLIPでは性能が低い.

  次に, 最適化を用いた\textbf{OPT}について述べる.
  \textbf{OPT}は, ほとんどのケースで\textbf{ALL}よりも高い性能が出ている.
  \textbf{ALL}では難しかったLightやDisplay, Waterについても, VQA(OFA)やITR(ImageBind)では100\%近い正答率を達成している.
  多様なドアを同じテキストのみで扱うVarious Doorsについても, 90\%以上の正答率を達成している.
  なお, 一つのテキストのみ用いる\textbf{ONE}は, \textbf{ALL}に比べると性能が良い一方で, \textbf{OPT}よりも性能は低い.

  最後に, 全体的な傾向について述べる.
  正答率の平均を見ると, VQA(OFA)とITR(ImageBind)は同程度の正答率であり, 最適化を行う\textbf{OPT}では95\%程度, 最適化を行わない\textbf{ALL}でも80\%以上の性能を叩き出している.
  その一方で, VQA(BLIP2)とITR(CLIP)は\textbf{ALL}が60\%程度とその性能は大きく落ち, 特にITR(CLIP)の性能は\textbf{OPT}でもVQA(OFA)やITR(ImageBind)の\textbf{ALL}と同程度の性能しか出ていない.
  正答率の分散を見ると, どのモデルについても, 最適化を行うことで, 認識する状態ごとの正答率の分散が小さくなっていることがわかる.
  その中でも特に, ITR(ImageBind)における\textbf{OPT}の正答率の分散は他のモデルに比べて非常に小さい.
}%

\begin{landscape}
\begin{table}[htb]
  \centering
  \caption{The result of the state recognition experiment. The percentage of correct responses is shown for four different models.}
  \begin{tabular}{|l|l||c|c|c||c|c|c||c|c|c||c|c|c|} \hline
    \multicolumn{2}{|c||}{Model} & \multicolumn{3}{|c||}{VQA(BLIP2)} & \multicolumn{3}{|c||}{VQA(OFA)} & \multicolumn{3}{|c||}{ITR(CLIP)} & \multicolumn{3}{|c|}{ITR(ImageBind)}\\ \hline
    \multicolumn{2}{|c||}{Method} & OPT & ONE & ALL & OPT & ONE & ALL & OPT & ONE & ALL & OPT & ONE & ALL \\ \hline\hline
    \multirow{2}{*}{Room} & $D_{opt}$               & 100 & 100 &  90  & 100 & 100 & 100  &  85 &  60 &  50  &  95 &  90 &  75\\ \cline{2-2}
                          & $D_{eval}$              & 100 & 100 &  80  & 100 & 100 & 100  &  70 &  55 &  50  &  90 &  90 &  90\\ \hline
    \multirow{2}{*}{Elevator} & $D_{opt}$           &  90 &  75 &  65  & 100 & 100 &  95  & 100 & 100 & 100  & 100 & 100 &  85\\ \cline{2-2}
                              & $D_{eval}$          &  85 &  60 &  65  & 100 & 100 & 100  & 100 & 100 & 100  &  90 & 100 &  85\\ \hline
    \multirow{2}{*}{Cabinet} & $D_{opt}$            & 100 & 100 &  35  & 100 & 100 & 100  & 100 &  95 &  65  & 100 & 100 & 100\\ \cline{2-2}
                             & $D_{eval}$           &  95 &  95 &  45  & 100 & 100 & 100  & 100 &  90 &  45  &  95 &  95 & 100\\ \hline
    \multirow{2}{*}{Refrigerator} & $D_{opt}$       & 100 & 100 &  80  & 100 & 100 &  60  & 100 & 100 & 100  & 100 & 100 & 100\\ \cline{2-2}
                                  & $D_{eval}$      & 100 & 100 &  70  & 100 & 100 &  70  & 100 & 100 &  90  & 100 & 100 & 100\\ \hline
    \multirow{2}{*}{Microwave} & $D_{opt}$          & 100 & 100 &  40  &  85 &  80 &  55  &  90 &  85 &  50  & 100 &  95 &  85\\ \cline{2-2}
                               & $D_{eval}$         &  95 &  95 &  40  &  85 &  85 &  60  &  85 &  65 &  50  &  90 &  90 & 100\\ \hline
    \multirow{2}{*}{Various Doors} & $D_{opt}$      &  95 &  89 &  62  &  89 &  83 &  79  &  88 &  85 &  77  &  98 &  91 &  81\\ \cline{2-2}
                             & $D_{eval}$           &  93 &  89 &  55  &  87 &  79 &  88  &  81 &  86 &  71  &  92 &  90 &  79\\ \hline
    \multirow{2}{*}{Transparent Door} & $D_{opt}$   &  80 &  80 &  55  &  90 &  80 &  50  &  70 &  65 &  50  &  90 &  70 &  45\\ \cline{2-2}
                                 & $D_{eval}$       &  75 &  75 &  50  &  70 &  70 &  50  &  55 &  45 &  45  &  80 &  60 &  40\\ \hline
    \multirow{2}{*}{Light} & $D_{opt}$              & 100 &  85 &  50  &  80 &  75 &  65  &  95 &  85 &  50  & 100 &  55 &  50\\ \cline{2-2}
                           & $D_{eval}$             &  90 &  50 &  50  &  75 &  65 &  60  &  85 &  90 &  50  &  90 &  50 &  50\\ \hline
    \multirow{2}{*}{Display} & $D_{opt}$            & 100 & 100 &  20  & 100 & 100 &  85  & 100 & 100 &  50  & 100 &  90 &  80\\ \cline{2-2}
                             & $D_{eval}$           &  95 &  95 &  35  &  95 &  95 &  75  &  95 &  90 &  50  &  90 &  90 &  90\\ \hline
    \multirow{2}{*}{Handbag} & $D_{opt}$            &  90 &  70 &  50  &  90 &  90 &  70  &  95 &  80 &  50  & 100 &  95 &  90\\ \cline{2-2}
                             & $D_{eval}$           &  60 &  80 &  50  &  90 &  90 &  80  &  85 &  65 &  50  &  85 &  85 &  90\\ \hline
    \multirow{2}{*}{Water} & $D_{opt}$              &  85 &  80 &  40  & 100 & 100 &  75  &  80 &  50 &  50  & 100 & 100 &  85\\ \cline{2-2}
                           & $D_{eval}$             &  85 &  85 &  30  &  85 &  85 &  75  &  60 &  50 &  50  &  90 &  90 &  85\\ \hline
    \multirow{2}{*}{Kitchen} & $D_{opt}$            &  95 &  85 &  50  & 100 & 100 & 100  &  95 &  90 &  70  & 100 & 100 & 100\\ \cline{2-2}
                             & $D_{eval}$           &  85 &  85 &  50  & 100 & 100 & 100  &  95 &  90 &  65  & 100 & 100 & 100\\ \hline \hline
    \multirow{2}{*}{Average} & $D_{opt}$            &  95.3 &  89.9 &  57.8  & 95.6 & 93.9 & 82.3  &  90.9 &  80.3 &  62.1  & 98.2 & 91.1 & 81.7\\ \cline{2-2}
                             & $D_{eval}$           &  89.5 &  86.3 &  55.3  & 92.5 & 91.3 & 83.9  &  83.1 &  75.1 &  58.7  & 91.5 & 88.0 & 85.9\\ \hline \hline
    \multirow{2}{*}{Standard Deviation} & $D_{opt}$ &  6.2  &  10.3 &  20.8  & 6.6  & 9.1  & 18.0  &  8.4  &  16.4 &  17.4  & 2.9  & 12.3 & 16.2\\ \cline{2-2}
                             & $D_{eval}$           &  10.7 &  14.5 &  15.6  & 9.7  & 11.5 & 17.3  &  14.5 &  19.0 &  16.1  & 5.3  & 13.9 & 17.4\\ \hline
  \end{tabular}
  \label{table:experiment}
\end{table}
\end{landscape}

\subsection{Navigation Experiment}
\begin{figure}[htb]
  \centering
  \includegraphics[width=0.8\columnwidth]{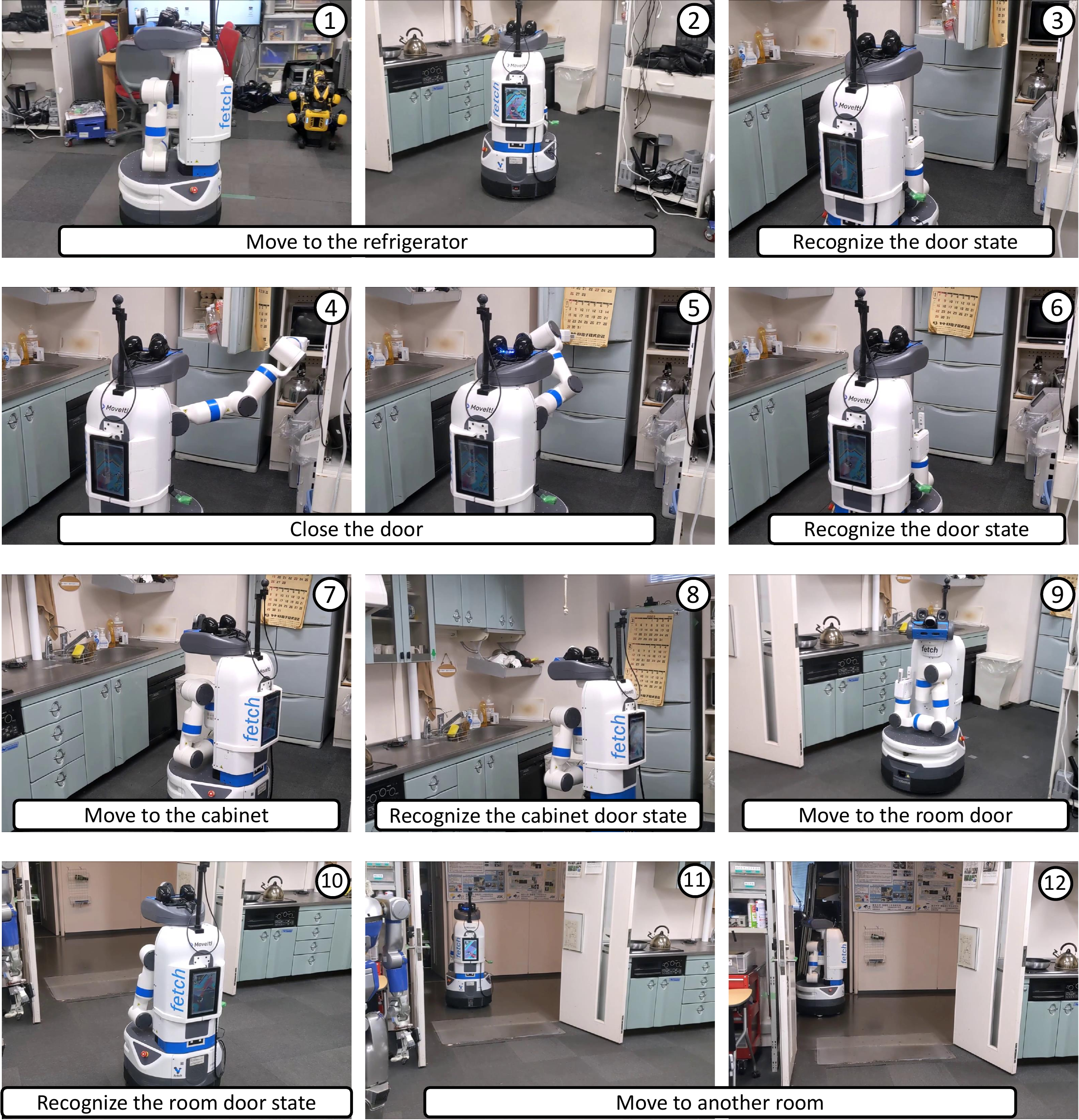}
  \caption{Navigation experiment including recognition of the refrigerator door state, cabinet door state, and room door state.}
  \label{figure:advanced}
\end{figure}

\switchlanguage%
{%
  We integrate our state recognition method and the mobile robot Fetch into a system, construct a practical application example, and verify the effectiveness of our method.
  Fetch recognizes the open/closed state of the refrigerator door in the kitchen, the open/closed state of the cabinet, and the open/closed state of the room door, in that order.
  If the refrigerator door is open, the robot closes it, if the cabinet is open, the robot closes it, and if the room door is open, the robot exits the room.
  Note that since ITR(ImageBind) with \textbf{ALL} is used for the recognition in this experiment, no optimization is required, and the same $Q$ as in \figref{figure:experiment} is used.
  \textbf{ALL} has the advantage of not requiring any data collection at all.
  Additionally, if navigation is possible with \textbf{ALL}, better performance can be achieved by using \textbf{OPT} or \textbf{ONE}.
  The robot positions for recognizing a refrigerator, a cabinet, and a room door, as well as the door-closing motion, are prepared in advance.
  The actual experiment is shown in \figref{figure:advanced}.
  First, the robot recognized that the refrigerator door is left open at \ctext{3}, closed the door by hand, and recognized that the door is closed at \ctext{6}.
  Next, the robot confirmed that the cabinet door is closed at \ctext{8}, and then moved on to the next action.
  Finally, the robot recognized that the room door is open at \ctext{10}, and left the room.
}%
{%
  本研究の状態認識手法と台車型ロボットFetchをシステム統合し, 実応用例を構築, その有効性を検証する.
  キッチンの冷蔵庫ドアの開閉状態, キッチン棚の開閉状態, 部屋の外に繋がるドアの開閉状態を順に認識する.
  冷蔵庫のドアが開いていたら閉め, キッチン棚が開いていたらそれを止め, ドアが開いていれば部屋の外に出る.
  なお, 本実験の認識にはITR(ImageBind)における\textbf{ALL}のみを用いるため, 最適化等は必要なく, \figref{figure:experiment}と同じ$Q$を利用している.
  \textbf{ALL}は一切のデータ収集が必要ない利点があり, また, この\textbf{ALL}でもnavigationができれば, \textbf{OPT}や\textbf{ONE}を用いることでより良い性能を得ることができる.
  また, 冷蔵庫やキッチン棚, 通常のドアを認識する位置やドアを閉める動作は予め準備されている.
  実際の実験の様子を\figref{figure:advanced}に示す.
  まず, \ctext{3}で冷蔵庫のドアが開きっぱなしであることを認識し, 手でドアを閉め, \ctext{6}でドアが閉まったことを認識した.
  次に, \ctext{8}でキッチン棚のドアが開いていないことを確認し, 次の動作に移った.
  最後に, \ctext{10}でドアが開いていることを認識し, 部屋の外に移動するという一連の動作に成功している.
}%

\section{Discussion} \label{sec:discussion}
\switchlanguage%
{%
  We discuss the experimental results of this study.
  From the experiments, it was found that various states can be recognized without any optimization depending on the model.
  For example, if the robot wants to recognize whether a door is open or closed, it only needs to express the state using the spoken language.
  On the other hand, some models can be inaccurate without optimization.
  In this study, the performance of VQA(OFA) and ITR(ImageBind) was higher than that of VQA(BLIP2) and ITR(CLIP).
  It is highly likely that generalization performance of state recognition can be ensured by using models trained with multiple tasks and modalities rather than a single task or a few modalities.
  We also found that the optimization procedure improves the performance of state recognition for all of the models.
  Especially for VQA(OFA) and ITR(ImageBind), the optimization achieved more than 90\% correct responses in most cases, indicating that sufficient recognition performance can be achieved by simply adjusting the weighting for each text, without manual programming or training of neural networks.
  It is expected that state recognition will become easier to perform, and the source code and computer resource for state recognition will become easier to manage.
  Additionally, when multiple prompts are weighted through optimization, performance improves compared to selecting a single prompt through optimization.
  When selecting a single prompt through optimization, performance improves compared to using multiple prompts with equal weights.
  This indicates that due to significant performance differences among individual prompts, it is effective to use only those prompts with good performance rather than all of them.
  The state recognition in this study includes not only the open/closed state of room doors, but also the open/closed state of various doors including transparent doors, the on/off state of lights and displays, whether water is running or not, and the cleanliness of the kitchen.
  In particular, recognizing the open/closed state of transparent doors and the presence/absence of running water are difficult for manual programming based on depth sensors, and it is important that these states can be recognized only by image and language.
  The common sense acquired by VLMs also enabled the robot to recognize the qualitative cleanliness of the kitchen.
  It is also important to note that the same text prompt can be used to recognize the open/closed state of doors of room, refrigerator, microwave oven, elevator, and cabinet doords, and that it is not necessary to change the text if the states to be recognized are similar in nature.

  We discuss the limitations and future prospects of this study.
  First, we describe the range of applicability of our state recognition.
  Currently, we have found that various types of state recognition are possible, but we have not yet achieved perfect state recognition.
  In particular, recognition of transparent doors is difficult, and the correctness rate when using certain models is quite low.
  However, there is a possibility to improve the accuracy by using not only images but also other modalities such as video, sound, and heatmaps.
  It is important to note that the reflection of light on transparent doors and water changes depending on the angle of view and time, so the use of video may improve the recognition performance.
  In addition, the use of multiple models at the same time will enable more accurate state recognition.
  For example, by using the four models treated in this study simultaneously, it is possible to improve the accuracy by selecting the best model for each state recognition, compensating for the disadvantages of each model.
  On the other hand, increasing the number of models leads to problems such as longer inference time and difficulties in resource management.
  On a different note, not only binary recognition like in this study, but also more advanced continuous state recognition from when the door opens until it closes is intriguing \cite{kawaharazuka2024clipseq}.
  We will carefully monitor the future development of the underlying models to realize a simpler, resource-manageable, and more accurate state recognition.
  Next, we discuss the process of preparing the text $Q$.
  In this study, we generated multiple texts by changing articles, state expressions, words, and question/expression forms.
  On the other hand, it is desirable that the state recognizer is automatically generated from only the linguistic command.
  By using a large-scale language model (LLM) such as GPT-4 \cite{achiam2023gpt4}, it is possible to automatically generate a variety of texts with the same meaning but with different expressions \cite{kanazawa2023cooking}.
  If such a mechanism can be introduced to improve resource management and performance, it will be possible to construct a more practical recognition system.
}%
{%
  本研究の実験結果について考察する.
  実験から, モデルによっては, 一切の最適化を行わなくても, 多様な状態認識が可能であることがわかった.
  例えばドア開閉を認識したければ, ドアが開いている状態と開いていない状態を言語で表現するだけで, 状態認識が可能である.
  一方で, モデルによっては, 最適化なしでは著しく精度が悪い場合がある.
  本研究では, OFAとImageBindの性能がBLIP2やCLIPを用いた場合よりも高かったことから, 単一タスクや少数モーダルではなく, 複数のタスクやモーダルを用いて学習を行ったモデルを用いることで, 状態認識の汎化性能を確保することができる可能性が高い.
  また, 最適化を行うことで, どのモデルでも性能が向上することがわかった.
  特にVQA(OFA)やITR(ImageBind)では, 最適化によりほとんどのケースで90\%以上の正答率を達成しており, 手動のプログラミングやニューラルネットワークの再学習等をせずとも, 各テキストに対する重みの調整のみで, 十分な認識性能を達成可能であることがわかった.
  今後, より簡単に状態認識を行うことができ, かつそのリソースやコードの管理も容易になることが期待される.
  加えて, 最適化によって複数のプロンプトを重み付けした場合は, 最適化によって一つのプロンプトを選ぶ場合よりも性能が向上する.
  最適化によって一つのプロンプトを選ぶ場合は, 複数のプロンプトを同じ重みで用いた場合よりも性能が向上する.
  これは, 個々のプロンプトの性能差が大きいため, 全てを使うのではなく, 一部の良い性能を持ったプロンプトのみを使うことが有効であることを示している.
  本研究の状態認識には, 通常のドアの開閉だけでなく, 透明なドアを含む多様なドアの開閉, 電気やディスプレイのオンオフ, 出水の有無, キッチンの綺麗さ等, 様々なケースを含む.
  特に, 透明なドアの開閉や出水の有無は, 深度センサに基づく手動のプログラミングでは難しいケースであり, これが言語のみで認識可能であることは重要である.
  キッチンの綺麗さについても, 大規模視覚-言語モデルが獲得した一般知識・常識により, 質的な状態認識が可能であった.
  また, 同じテキストプロンプトで, 通常のドアや冷蔵庫, 電子レンジ, エレベータ, 棚のドア開閉を認識可能であり, 似た性質のものであればテキストを変更する必要すらないことも重要な点である.

  本研究の限界と今後の展望について述べる.
  まず, 本状態認識が適用可能な範囲の拡大について述べる.
  現状, 様々な状態認識が可能であることがわかったが, まだ完璧な状態認識が可能となったわけではない.
  特に透明なドアの認識は難しく, モデルによっては正答率がかなり低い場合が存在する.
  これに対して, 画像だけでなく, 動画や音, ヒートマップのような別のモダリティを同時に利用することで, その精度を向上させることができる可能性がある.
  透明なドアや水は, 画角や時間で光の反射が変化する点は重要であり, 動画を用いることで認識性能が向上する可能性がある.
  また, 複数のモデルを同時に利用することで, より高精度な状態認識が可能となる.
  少なくとも, 本研究で扱った4つのモデルを同時に利用することで, 各モデルの不得意を補い合い, 各状態認識について最も良いモデルを選択し精度を向上させることができる.
  一方で, モデルを増やすことで, 推論時間が長くなる, リソースの管理が難しくなるといった問題も発生する.
  また, 本研究のような2値の分類だけでなく, より高度な, ドアが開いてから閉まるまでの連続的な状態認識も興味深い\cite{kawaharazuka2024clipseq}.
  今後の基盤モデルの発展を注意深く観察し, より簡易かつリソース管理が容易であり, 高精度な状態認識を実現していきたい.
  次に, テキスト$Q$の選択方法について述べる.
  本研究ではテキストを, 冠詞や状態表現, 単語, 質問/表現形式の種類を変更することで複数生成した.
  一方で, ドアの開閉を認識したいという言語指令のみから, 状態認識器が自動生成されることが望ましい.
  GPT-3 \cite{brown2020gpt3}のような大規模言語モデルを利用することで, 表現を変えた様々なテキストを自動生成することが可能である\cite{kanazawa2023cooking}.
  このような仕組みを導入し, リソース管理や性能を向上させることができれば, より実用的な認識システムの構築が可能となる.
}%

\section{Conclusion} \label{sec:conclusion}
\switchlanguage%
{%
  In this study, we proposed an environment state recognition method for robots using pre-trained large-scale vision-language models (VLMs).
  By applying two tasks of VLMs, Visual Question Answering (VQA) or Image-to-Text Retrieval (ITR), the robot can recognize the open/closed state of room doors and the on/off state of lights by simply preparing multiple texts that represent the state to be recognized.
  We have also shown that the recognition accuracy can be improved by selecting appropriate texts from the set of prepared texts based on black-box optimization, and that it is possible to recognize various states including the open/closed state of transparent doors, whether water is running or not, and even the cleanliness of a kitchen.
  We clarified the performance difference among the models, and the strengths and weaknesses of each model with regards to its recognizable states.
  Since this study does not require training of neural networks or manual programming, there is no need to prepare several different models and programs, and the source code and computer resource can be easily managed.
  In the future, we would like to study multi-modalization of recognizers, automatic text generation, and automatic model selection, in order to construct more practical robot systems.
}%
{%
  本研究では, 事前学習済みの大規模視覚-言語モデルを用いたロボットにおける環境状態認識を提案した.
  大規模視覚-言語モデルのタスクであるVisual Question Answering (VQA)とImage-to-Text Retrieval (ITR)を応用し, 認識したい状態を表現するテキストを複数用意するのみで, ドアの開閉認識や電気のオンオフ認識が可能となる.
  また, 用意した多数のテキスト集合の中からブラックボックス最適化に基づき適切なもの選択することで, よりその認識精度を向上させることが可能であり, 透明なドアの開閉や水の認識, キッチンの綺麗さまでも含む, 多様な状態認識が可能となることを示した.
  その認識性能はモデルごとに大きく異なり, 各モデルにおける性質, 認識可能な状態の得意不得意を明らかにした.
  本研究はニューラルネットワークの再学習や手動のプログラミング等を行わないため, 異なるモデルやプログラムを複数用意する必要がなく, コードやリソースの管理が容易になる.
  今後, より実用的なロボットシステム構築を目指し, 認識器のマルチモーダル化や, テキストの自動生成, モデルの自動選択等について研究を行っていきたい.
}%

{
  \bibliographystyle{junsrt}
  \bibliography{main}
}

\end{document}